\documentclass{article}
\usepackage{ijcai26}

\usepackage{times}
\usepackage{url}
\usepackage[hidelinks]{hyperref}
\usepackage[utf8]{inputenc}
\usepackage[small]{caption}
\usepackage{graphicx}
\usepackage{amsmath}
\usepackage{amssymb}
\usepackage{booktabs}
\usepackage{tabularx}
\usepackage{makecell}
\usepackage{multirow}

\title{SeesawNet: Towards Non-stationary Time Series Forecasting with \\ Balanced Modeling of Common and Specific Dependencies}

\author{
    Hao Li$^1$\and
    Lu Zhang$^2$\thanks{Corresponding authors.}\and
    Liu Chong$^1$\and
    Yankai Chen$^3$\and
    Pengyang Wang$^4$\And
    Yingjie Zhou$^1$\footnotemark[1]\\
    \affiliations
    $^1$Sichuan University\\
    $^2$Chengdu University of Information Technology\\
    $^3$McGill University\\
    $^4$University of Macau\\
    \emails
    haonlee98@stu.scu.edu.cn,
    zhanglu@cuit.edu.cn,
    yihann@stu.scu.edu.cn,
    yankaichen@acm.org,
    pywang@um.edu.mo,
    yjzhou09@gmail.com
}

\begin{document}

\maketitle

\begin{abstract}
    Instance normalization (IN) is widely used in non-stationary multivariate time series forecasting to reduce distribution shifts and highlight common patterns across samples. However, IN can over-smooth instance-specific structural information that is essential for modeling temporal and cross-channel heterogeneity.
    While prior methods further suppress distribution discrepancies or attempt to recover temporal specific dependencies, they often ignore a central tension: how to adaptively model common and instance-specific dependency based on each instance's non-stationary structures. To address this dilemma, we propose {\it SeesawNet}, a unified architecture that dynamically balances common and instance-specific dependency modeling in both temporal and channel dimensions. At its core is {\it Adaptive Stationary–Nonstationary Attention (ASNA)}, which captures common dependencies from normalized sequences and specific dependencies from raw sequences, and adaptively fuses them according to instance-level non-stationarity. Built upon ASNA, SeesawNet alternates dedicated temporal and channel relationship modeling to jointly capture long-range and cross-variable dependencies. Extensive experiments on multiple real-world benchmarks demonstrate that SeesawNet consistently outperforms state-of-the-art methods.
\end{abstract}

\noindent\textbf{Code:} \url{https://github.com/dreamone-Lee/SeesawNet}

\section{Introduction}

    \begin{figure}[t]
        \centering
        \includegraphics[width= 0.99 \linewidth]{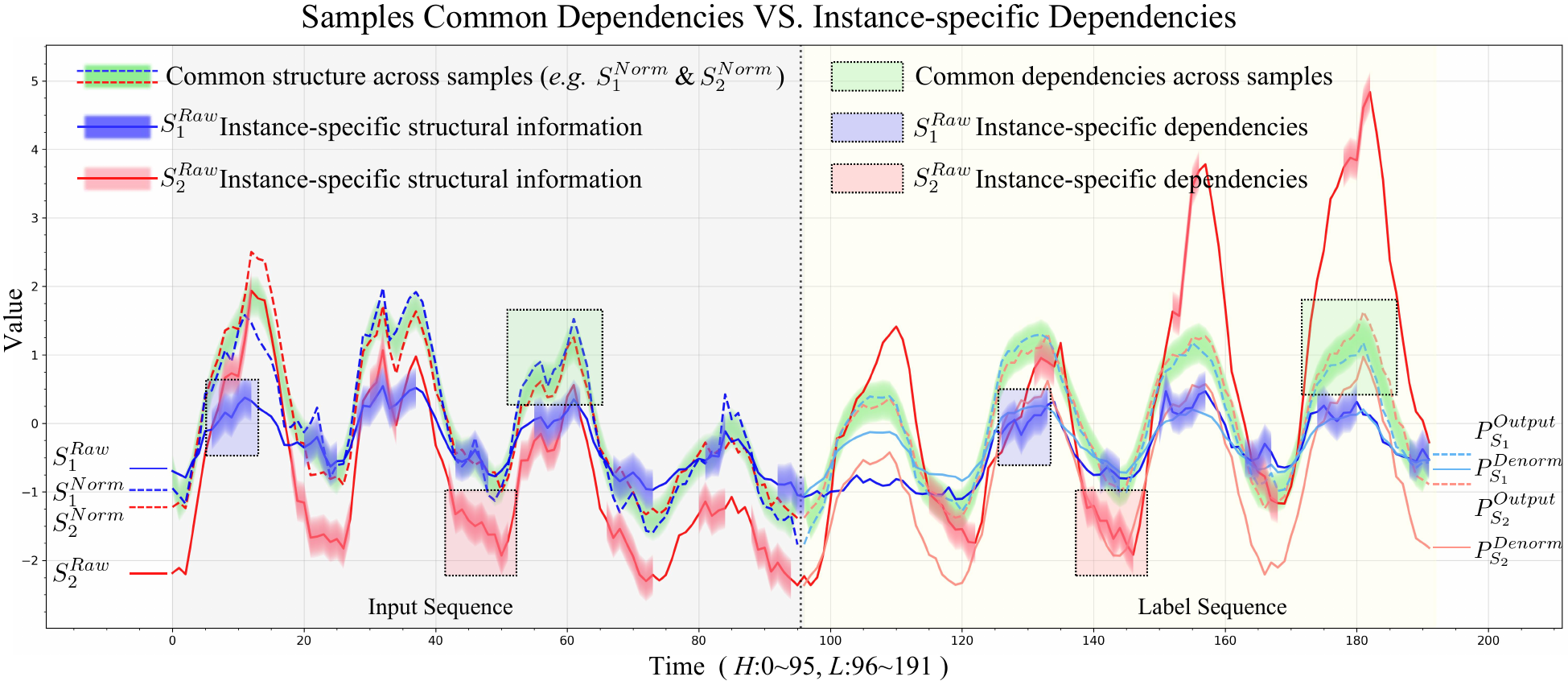}
        \caption{
            Illustration of our motivation from an empirical example.
            Two raw sequences $S_1^{\mathrm{Raw}}$ and $S_2^{\mathrm{Raw}}$ exhibit substantially different distributions. After IN, their normalized inputs $S_1^{\mathrm{Norm}}$ and $S_2^{\mathrm{Norm}}$ reveal clearer cross-sample common structures, yielding similar forecasts $P_{S_1}^{\mathrm{Output}}$ and $P_{S_2}^{\mathrm{Output}}$, indicating that transferable common dependencies are well captured.
            However, sample-specific trends and local fluctuations in the raw sequences are not sufficiently reflected in the predictions, implying that IN may attenuate instance-specific dependencies. This example highlights the complementary roles and inherent limitations of both in non-stationary forecasting.
            }
        \label{fig:main}
    \end{figure}

    Multivariate time series forecasting (MTSF) plays a crucial role in a wide range of real-world applications, such as traffic management \cite{zhang2023crossformer}, power load prediction \cite{cao2025dense}, financial investment \cite{lin2025cspo}, and weather forecasting \cite{liu2023itransformer}.
    However, real-world time series often exhibit strong non-stationarity, characterized by continuously shifting distributions over time due to factors such as trends, seasonality, and irregular events \cite{ma2024u,kim2021reversible}.
    Such non-stationarity induces substantial distributional disparities across samples, which
    can markedly limit the model's learning capacity and pose challenges to generalization \cite{kim2025battling,liu2022non}.

    To mitigate non-stationarity, instance normalization (IN) and its variants have become a common design choice in modern forecasting models \cite{qiu2025duet,chen2024pathformer,wang2024timemixer++}.
    A representative paradigm is RevIN \cite{kim2021reversible}, which normalizes each input instance before modeling and restores it afterward, effectively aligning distributions and facilitating the learning of common patterns.
    Recent variants further refine normalization/denormalization to ease learning under severe non-stationarity \cite{dai2024ddn,ye2024frequency,han2024sin}.
    However, in aggressively suppressing non-stationary components, instance normalization also alters instance-level statistics, which may obscure genuine structural information (e.g., instance-specific trends, periodicities, or unique patterns), thereby weakening the modeling of temporal and cross-channel heterogeneity.
    As illustrated in Fig.~\ref{fig:main}, two samples with substantially different raw distributions may become highly similar after IN, leading to similar forecasts while their instance-specific structural information are diminished.

    In power load forecasting, seasonal shifts and usage behaviors induce distinct temporal structures critical for accurate prediction \cite{cao2025dense}.
    After IN, instance-level temporal cues may be attenuated, biasing the model toward shared dynamics and reducing sensitivity to instance-specific evolutions.
    Several methods (e.g., NS-Transformer \cite{liu2022non} and U-Mixer \cite{ma2024u}) attempt to recover instance-specific temporal dependencies via attention correction or correlation-based constraints.
    Yet, since instances may exhibit different degrees of non-stationarity, treating all samples uniformly as purely “common-pattern” or purely “instance-specific” cases can be suboptimal; accordingly, existing methods generally lack an instance-adaptive coordination between common-dependency learning and instance-specific temporal modeling.

    \begin{figure}[t]
        \centering
        \includegraphics[width= 0.99 \linewidth]{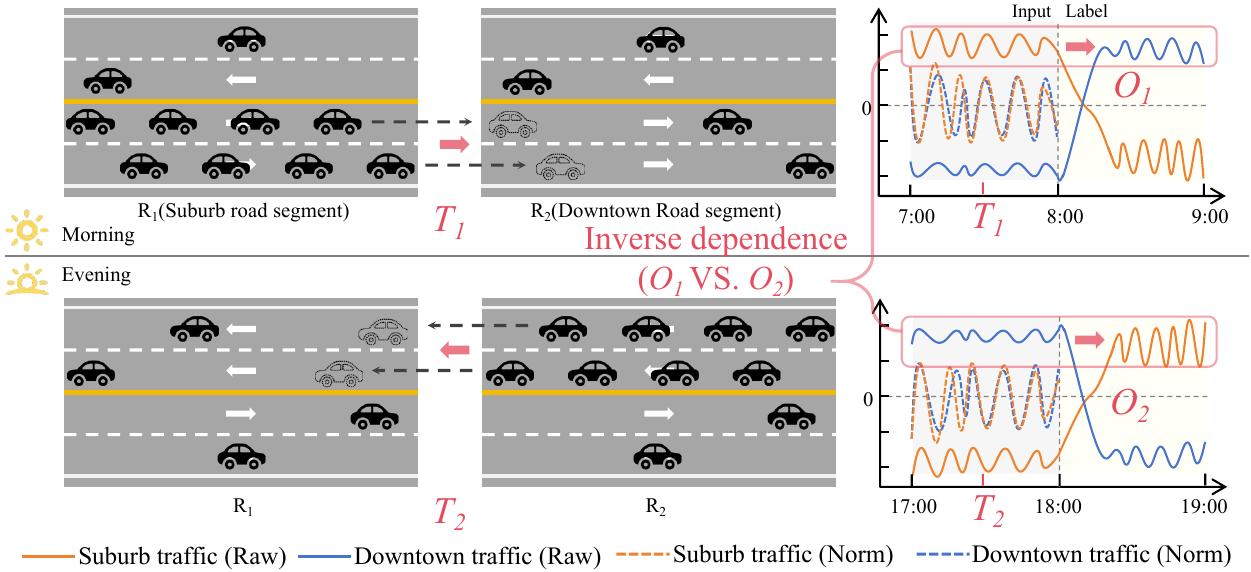}
        \caption{
            The observation of blurred regime-dependent cross-channel dependencies caused by instance normalization (IN).
        In traffic forecasting, the relation between $R_1$ and $R_2$ may reverse across regimes ($O_1$ VS. $O_2$).
        As IN suppresses regime-dependent structural cues (e.g., flow magnitudes and scale differences), these inverse dependency patterns become less distinguishable, making such instance-specific cross-channel dependencies harder to capture.
        }
        \label{fig:traffic}
    \end{figure}

    Beyond temporal dynamics, IN may also hinder cross-channel dependency modeling that are vital for multivariate forecasting.
    For example, in traffic flow prediction, the interdependence between correlated road segments may exhibit divergent dynamic coordination patterns \cite{zhang2023crossformer}.
    As shown in~Fig.~\ref{fig:traffic},  the reversal is evident in raw signals, yet becomes much less distinguishable after IN, weakening instance-specific cross-channel dependencies.
    Despite its practical significance, this challenge remains underexplored in existing literature.
    Therefore, we ask: \emph{how can we adaptively model common and instance-specific dependency, guided by each instance’s non-stationary structures, in both temporal and channel dimensions?}

    To address the above challenges, we propose \emph{SeesawNet}, a unified forecasting framework that dynamically balances the modeling of common and instance-specific dependencies under non-stationarity.
    At its core is \emph{Adaptive Stationary-Nonstationary Attention (ASNA)}, which uses two complementary attention branches to extract common dependencies from normalized sequences and instance-specific dependencies from raw sequences, and fuses them via an instance-adaptive gating mechanism.
    Built upon ASNA, SeesawNet further introduces a \emph{Patch Dependency Learning Layer} to model temporal balanced dependencies within each channel across time intervals, and a \emph{Channel Relationship Learning Layer} to model shared and patch-specific cross-channel dependencies at each temporal patch.

    By jointly considering the balance between common and specific dependencies along both temporal and channel axes, SeesawNet avoids the bias of IN-based models toward overly common patterns while preserving instance-specific structures.
    Our contributions are summarized as follows:
    \begin{itemize}
    \item
        We identify an instance-level dilemma in non-stationary forecasting: raw sequences contain rich structures but exhibit large distributional discrepancies, whereas instance normalization reduces such discrepancies while blurring instance-specific dependencies across both temporal and cross-channel dimensions.
    \item
        We propose SeesawNet, a unified ASNA-based framework that is not a simple module stacking but an instance-adaptive mechanism for coordinating common and instance-specific dependencies across temporal and cross-channel dimensions.

    \item
        Extensive experiments on multiple real-world benchmarks show that SeesawNet consistently outperforms state-of-the-art methods, and further evaluations demonstrate that ASNA generalizes across different Transformer-based forecasting models.
\end{itemize}

\section{Related Work}
    \subsection{Instance Normalization (IN) in Multivariate Time Series Forecasting}

        To mitigate distributional discrepancies induced by non-stationarity, IN has become a widely adopted technique in time series forecasting.
        Representative methods such as RevIN~\cite{kim2021reversible} normalize each input instance to stabilize input distributions, facilitating the learning of common patterns, and have been broadly integrated into model-agnostic forecasting frameworks, including channel-independent methods \cite{dai2024periodicity,nie2022time,xu2023fits} and channel-dependent approaches \cite{qiu2025duet,chen2024pathformer,liu2023itransformer}.

        To further reduce non-stationarity, follow-up research refines the normalization process from three major perspectives: enhancing representational ability, increasing granularity, and improving de-normalization accuracy.
        For instance, some studies introduce richer time- or frequency-domain statistical features \cite{ye2024frequency,han2024sin} to better characterize non-stationary behaviors.
        Others employ finer-grained normalization schemes, such as slice-wise or point-wise normalization \cite{dai2024ddn,liu2023adaptive}, to improve adaptability to varying temporal non-stationarities.  Subsequent works enhance the flexibility of the de-normalization stage by optimizing output-level parameters \cite{fan2024deep,fan2023dish}.
        Overall, these variants primarily aim at improving distribution alignment or restoring scales more accurately; meanwhile, how IN affects instance-specific dependency modeling in temporal and channel dimensions is less explicitly discussed.

    \subsection{Learning Dependencies after Normalization}
        Since IN may attenuate instance-specific signals, recent research has attempted to recover or enhance instance-specific dependencies within normalized sequences.
        NS-Transformer \cite{liu2022non} studies the problem of over-stationarization and suggests that IN can reduce sample distinctions in the attention space, which may weaken temporal dependency modeling; accordingly, it uses raw input signals to correct attention scores after normalization.
        From the perspective of distributional evolution, U-Mixer \cite{ma2024u} enforces constraints to align correlation patterns between inputs and outputs, aiming to preserve sample-wise temporal consistency.

        These efforts mainly focus on recovering temporal dependencies after normalization.
        For multivariate forecasting, cross-channel dependency patterns are also sensitive to normalization, and preserving instance-specific inter-variable relations remains relatively underexplored.
        In addition, existing methods often emphasize instance-specific cues through correction or constraints, while how to coordinate common dependency learning and instance-specific dependency modeling under IN remains an open question.
        A recent study, TimeBridge \cite{liu2024timebridge}, discusses balancing the elimination and retention of non-stationarity from a representation perspective, i.e., mitigating non-stationary components for short-term variations while maintaining them for long-term cointegration structures after IN.
        Its notion of ``balance'' is different from ours: we focus on dynamically coordinating common and instance-specific dependency modeling induced by IN across both temporal and channel dimensions.

    Despite notable progress in normalization strategies and post-normalization dependency learning, it remains unclear how to retain instance-specific dependency structures while preserving the benefits of IN for learning common dependencies in both temporal and channel dimensions.
    To fill this gap, we propose SeesawNet, which systematically models this trade-off from both temporal and channel perspectives.

\section{Methods}

    \subsection{Problem Formulation}
\begin{figure*}[!t]
        \centering
        \includegraphics[width= 0.95 \textwidth]{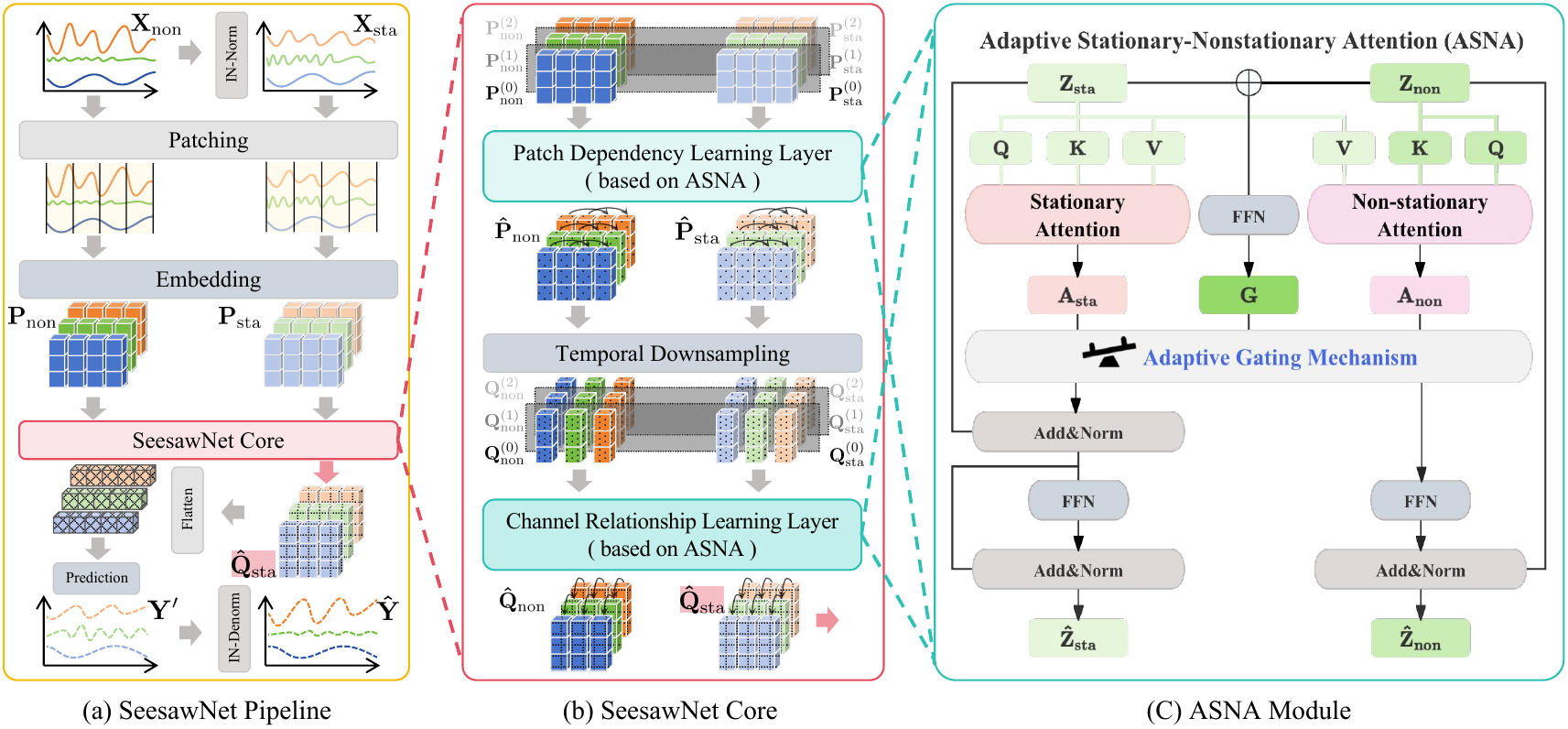}
        \caption{Overall architecture of SeesawNet with its core component Adaptive Stationary-non-stationary Attention (ASNA).}
        \label{fig:overall}
    \end{figure*}

        Given a multivariate time series instance with $C$ channels (variables), we denote the historical input as $\mathbf{X}\in \mathbb{R}^{C\times L}$ and the forecasting target as $\mathbf{Y}\in \mathbb{R}^{C\times H}$, where $L$ and $H$ are the input and prediction horizons, respectively.
        The goal of multivariate time series forecasting is to learn a mapping $f(\cdot)$ such that $\hat{\mathbf{Y}} = f(\mathbf{X})$ approximates $\mathbf{Y}$.
        In non-stationary scenarios, instance normalization (IN) is commonly used to reduce distribution shifts and facilitate learning common patterns, but it may also attenuate instance-specific structures that are important for temporal and cross-channel dependency modeling.
        Motivated by this trade-off, we propose SeesawNet (Fig.~\ref{fig:overall}), which balances common and instance-specific dependency modeling along both temporal and channel dimensions, with ASNA as the reusable building block in both dependency learning layers.
        Overall, SeesawNet builds normalized/raw paths, applies ASNA for temporal patch and cross-channel dependency learning, and flattens the balanced representation for prediction before de-normalization.

    \subsection{Patching \& Embedding}
        To preserve both general and specific information, we adopt a dual-path input strategy(Fig.~\ref{fig:overall}(a)), which differs from conventional normalized-only designs.
        Specifically, given $\mathbf{X}\in\mathbb{R}^{C\times L}$, we compute the stationary sequence $\mathbf{X}_{\text{sta}}=\mathrm{IN\text{-}Norm}(\mathbf{X})$ via instance normalization along the temporal axis for each channel (RevIN-style \cite{kim2021reversible}), and retain the original non-stationary input $\mathbf{X}_{\text{non}}=\mathbf{X}$, where $\mathbf{X}_{\text{sta}}, \mathbf{X}_{\text{non}}\in\mathbb{R}^{C\times L}$.

        Then, $\mathbf{X}_{\text{sta}}$ and $\mathbf{X}_{\text{non}}$ are partitioned into overlapping patches of length $P$ with stride $S$.
        Following common patching practice, we pad the tail when necessary to ensure the whole input horizon is covered, yielding $N$ patches per channel.
        The resulting patches are embedded into a $D$-dimensional space:
        \begin{equation}
        \mathbf{P}_{\text{sta}}=\mathrm{Embedding}(\mathrm{Patching}(\mathbf{X}_{\text{sta}})),
        \end{equation}
        \begin{equation}
        \mathbf{P}_{\text{non}}=\mathrm{Embedding}(\mathrm{Patching}(\mathbf{X}_{\text{non}})).
        \end{equation}

    \subsection{Adaptive Stationary-Nonstationary Attention}

        To address the issue where IN enhances learnability but suppresses instance-specific structural information, we design the ASNA module(Fig.~\ref{fig:overall}(c)), which consists of three components and serves as the seesaw between normalization and structural fidelity.

        ASNA is formulated as a reusable attention block operating on a generic token sequence.
        Given stationary and non-stationary token embeddings $\mathbf{Z}_{\text{sta}}$ and $\mathbf{Z}_{\text{non}}$ (e.g., derived from $\mathbf{P}_{\text{sta}}$ and $\mathbf{P}_{\text{non}}$), ASNA outputs updated embeddings $\mathbf{\hat{Z}}_{\text{sta}}$ and $\mathbf{\hat{Z}}_{\text{non}}$, where $\mathbf{Z}_{\text{sta}},\mathbf{Z}_{\text{non}}\in\mathbb{R}^{T\times D}$, $T$ is the token length and $D$ is the hidden dimension.
        Importantly, $\mathbf{Z}$ corresponds to different semantic tokens in the two dependency learning layers:
        (i) in the \emph{Patch Dependency Learning Layer}, ASNA models temporal dependencies within each channel by treating $\mathbf{Z}$ as the patch tokens of that channel, i.e., $\mathbf{Z}\equiv \mathbf{P}^{(c)}$ with $T=N$;
        (ii) in the \emph{Channel Relationship Learning Layer}, ASNA models cross-channel dependencies within each patch by treating $\mathbf{Z}$ as the channel tokens at that patch, i.e., $\mathbf{Z}\equiv \mathbf{Q}^{(n)}$ with $T=C$.
        This abstraction allows ASNA to serve as the shared foundation for both temporal and cross-channel dependency modeling.

        \noindent\textbf{Stationary Attention.}
            Stationary embeddings are obtained from normalized sequences and thus exhibit reduced scale discrepancies.
            As a result, attention computed on $\mathbf{Z}_{\text{sta}}$ tends to emphasize patterns that are stable and transferable across samples (e.g., periodicity, trend-related cues, and recurring channel coordination patterns).
            Accordingly, the stationary attention scores are computed as:
            \begin{equation}
                \mathbf{A}_{\text{sta}} = \mathrm{softmax}\left(\frac{\mathbf{Z}_{\text{sta}} \mathbf{Z}_{\text{sta}} ^\top}{\sqrt{d}}\right).
            \end{equation}

        \noindent\textbf{Non-stationary Attention.}
            In contrast, non-stationary embeddings are derived from raw sequences and retain richer instance-specific structural cues.
            When $\mathbf{Z}_{\text{non}}$ is used to compute attention scores, the resulting interactions can highlight instance-dependent contexts, facilitating the modeling of instance-specific dependencies.
            Hence, the non-stationary attention scores are computed as:
            \begin{equation}
            \mathbf{A}_{\text{non}}  = \mathrm{softmax}\left(\frac{\mathbf{Z}_{\text{non}} \mathbf{Z}_{\text{non}} ^\top}{\sqrt{d}}\right).
            \end{equation}

        \noindent\textbf{Adaptive Gating Mechanism.}
            It generates a gate tensor $\mathbf{G}$ from concatenated representations, controlling the contribution of each attention branch:
            \begin{equation}
                \mathbf{G} = \sigma\left(\mathrm{Linear}\left(\mathbf{Z}_{\text{sta}} \oplus \mathbf{Z}_{\text{non}} \right)\right),
            \end{equation}
            where $\sigma(\cdot)$ is the sigmoid function.
            Note that $\mathrm{Linear}(\cdot)$ can be implemented equivalently by a token-wise $1\times1$ convolution without changing the underlying operation.

            The attention score formulations in Eq.(3)--(4) are simplified expressions of the score computation in multi-head attention.
            In our implementation, the fused attention output below is equivalent to combining two multi-head attention branches:
            (i) a stationary branch where $\mathbf{Q},\mathbf{K},\mathbf{V}$ are all projected from $\mathbf{Z}_{\text{sta}}$;
            (ii) a non-stationary branch where $\mathbf{Q},\mathbf{K}$ are projected from $\mathbf{Z}_{\text{non}}$, while $\mathbf{V}$ is projected from $\mathbf{Z}_{\text{sta}}$ to keep the aggregated content aligned with the stationary embedding space.
            Using the stationary embedding $\mathbf{Z}_{\text{sta}}$ as the attention value, we obtain the final fused attention output as follows:
            \begin{equation}
            \mathbf{O} = \left((\mathbf{1} - \mathbf{G}) \odot \mathbf{A}_{\text{sta}}  + \mathbf{G} \odot \mathbf{A}_{\text{non}}\right)\mathbf{Z}_{\text{sta}} ,
            \end{equation}
            where $\mathbf{O} \in \mathbb{R}^{T \times D}$ denotes the fused output.
            By combining stationary scores that emphasize common dependencies and non-stationary scores that highlight instance-specific dependencies, ASNA adaptively balances common and instance-specific dependency modeling according to the non-stationary properties of each instance.

            To enhance the model’s learning capacity, we follow the standard Transformer design with residual connections and feed-forward networks to update both embeddings.
            The fused output $\mathbf{O}$ serves as a shared interaction signal, while each branch preserves its own residual pathway:
            \begin{equation}
                \mathbf{Z}_{\text{tmp}} = \mathrm{Norm}(\mathbf{Z}_{\text{sta}}  + \mathrm{Dropout}(\mathbf{O} )) ,
            \end{equation}
            \begin{equation}
                \mathbf{\hat{Z}}_{\text{sta}} = \mathrm{Norm}(\mathbf{Z}_{\text{tmp}}  + \mathrm{FFN}(\mathbf{Z}_{\text{tmp}} )) ,
            \end{equation}
            \begin{equation}
                \mathbf{\hat{Z}}_{\text{non}} = \mathrm{Norm}(\mathbf{Z}_{\text{non}}  + \mathrm{FFN}(\mathbf{O} )),
            \end{equation}
            where ${\mathbf{Z}_{\text{tmp}}, \mathbf{\hat{Z}}_{\text{sta}}, \mathbf{\hat{Z}}_{\text{non}}} \in \mathbb{R}^{T \times D}$ denote the temporary, updated stationary, and updated non-stationary embeddings, respectively.

            These designs allow ASNA to coordinate common dependency learning and instance-specific dependency modeling within a unified attention framework.
            Notably, ASNA can substitute standard attention layers in Transformer-based models (e.g., iTransformer \cite{liu2023itransformer} and PatchTST \cite{nie2022time}), demonstrating broad applicability.

    \subsection{Patch Dependency Learning Layer}
        This layer focuses on modeling intra-channel temporal dependencies on the patch axis.
        For each channel $c\in[0,C-1]$, we treat its patch embeddings as the token sequence for ASNA, i.e., $\mathbf{Z}\equiv \mathbf{P}^{(c)}$ with token length $T=N$.
        Specifically, we feed the stationary patch embedding sequence $\mathbf{P}_{\text{sta}}^{(c)}=\mathbf{P}_{\text{sta}}[c,:,:]\in\mathbb{R}^{N\times D}$ and the non-stationary patch embedding sequence $\mathbf{P}_{\text{non}}^{(c)}=\mathbf{P}_{\text{non}}[c,:,:]\in\mathbb{R}^{N\times D}$ into ASNA, and obtain updated representations:
        \begin{equation}
            \mathbf{\hat{P}}_{\text{sta}}^{(c)}, \mathbf{\hat{P}}_{\text{non}}^{(c)} = \mathrm{ASNA}(\mathbf{P}_{\text{sta}}^{(c)}, \mathbf{P}_{\text{non}}^{(c)}).
        \end{equation}

        This enables each channel to capture temporal interactions across patches while balancing common dynamics and instance-specific evolutions.
        Stacking multiple layers further facilitates learning deeper temporal structures.

    \subsection{Channel Relationship Learning Layer}
        To capture inter-channel interactions at each temporal patch, we first apply a learnable temporal aggregation (implemented as $\mathrm{Conv1D}$ with kernel size 1) to compress the patch tokens and obtain a shorter sequence of $N'$ patches:
        \begin{equation}
        \mathbf{Q}_{\text{sta}}  = \mathrm{Conv1D}(\mathbf{\hat{P}}_{\text{sta}}),
        \end{equation}
        \begin{equation}
        \mathbf{Q}_{\text{non}}  = \mathrm{Conv1D}(\mathbf{\hat{P}}_{\text{non}}),
        \end{equation}
        where $\mathbf{Q}_{\text{sta}}, \mathbf{Q}_{\text{non}}\in\mathbb{R}^{C\times N'\times D}$ and $N'$ denotes the (aggregated) patch number.
        Then, for each temporal patch $n\in[0,N'-1]$, we treat the channel embeddings at this patch as the token sequence for ASNA, i.e., $\mathbf{Z}\equiv \mathbf{Q}^{(n)}$ with token length $T=C$.
        We feed $\mathbf{Q}_{\text{sta}}^{(n)}=\mathbf{Q}_{\text{sta}}[:,n,:]\in\mathbb{R}^{C\times D}$ and $\mathbf{Q}_{\text{non}}^{(n)}=\mathbf{Q}_{\text{non}}[:,n,:]\in\mathbb{R}^{C\times D}$ into ASNA to model cross-channel dependencies:
        \begin{equation}
        \mathbf{\hat{Q}}_{\text{sta}}^{(n)}, \mathbf{\hat{Q}}_{\text{non}}^{(n)} = \mathrm{ASNA}(\mathbf{Q}_{\text{sta}}^{(n)}, \mathbf{Q}_{\text{non}}^{(n)}),
        \end{equation}
        where $\mathbf{\hat{Q}}_{\text{sta}}^{(n)}, \mathbf{\hat{Q}}_{\text{non}}^{(n)}\in\mathbb{R}^{C\times D}$ represent the updated stationary and non-stationary embeddings at patch $n$.
        This layer enhances the model’s ability to capture cross-channel relations that may vary across different temporal regimes.

    \subsection{Flatten \& Prediction}
        Through patch-wise and channel-wise dependency learning layers, the updated stationary embedding $\mathbf{\hat{Q}}_{\text{sta}}$ encodes the balanced common and instance-specific dependencies across both temporal and channel dimensions.
        We use $\mathbf{\hat{Q}}_{\text{sta}}$ as the latent representation to predict the target sequence.
        Specifically, we flatten $\mathbf{\hat{Q}}_{\text{sta}}$ and apply a prediction head (denoted as $\mathrm{FFN}$, e.g., a lightweight linear/MLP head) to obtain:
        \begin{equation}
        \mathbf{Y'} = \mathrm{FFN}(\mathrm{Flatten}(\mathbf{\hat{Q}}_{\text{sta}})).
        \end{equation}
        Finally, we restore the original scale of the output via IN denormalization using the instance statistics from $\mathrm{IN\text{-}Norm}$:
        \begin{equation}
        \mathbf{\hat{Y}} = \mathrm{IN\text{-}Denorm}(\mathbf{Y'}).
        \end{equation}

\section{Experiments}

    \subsection{Experiment Setting}

        \noindent\textbf{Datasets.}
        We evaluate on nine real-world datasets: ETT (ETTh1/ETTh2 and ETTm1/ETTm2) \cite{wu2021autoformer}, Exchange Rate, Weather \cite{zeng2023transformers}, ILI, Solar-energy, and ECL \cite{liu2023itransformer}, covering 7 to 321 variables and frequencies from 10 minutes to 1 week. Dataset splits and horizons follow common settings in prior work.

        \noindent\textbf{Baselines.}
            We compare SeesawNet with representative models from \textbf{three categories}: \textbf{1)} Instance normalization-based improvements: DDN \cite{dai2024ddn}, SAN \cite{liu2023adaptive}; \textbf{2)} Dependency modeling after normalization: U-Mixer \cite{ma2024u}, NS-Transformer \cite{liu2022non}, TimeBridge \cite{liu2024timebridge}; and \textbf{3)} Time series forecasting models: PDF \cite{dai2024periodicity}, iTransformer \cite{liu2023itransformer}, PatchTST \cite{nie2022time}. These baselines cover normalization-based methods, post-normalization dependency modeling, and strong forecasting backbones.

            \noindent\textbf{Setups.} To ensure fair comparisons, we follow these settings: 1) Dataset splitting follows the strategy of iTransformer \cite{liu2023itransformer}; 2) Baselines use default optimal hyperparameters from their public implementations; 3) If settings are missing in the original work, we perform grid search to select the best configuration; 4) Input lengths follow model-specific recommendations. Prediction lengths are set to \{96, 192, 336, 720\}, except for ILI, where we use \{24, 36, 48, 60\}; and 5) We report MSE and MAE, averaged over three runs for each setting.

        \noindent\textbf{Implementation Details.} All experiments are implemented in PyTorch and run on two NVIDIA RTX 4090 GPUs (24GB). We adopt the Adam optimizer with learning rates chosen from \{1e-3, 2e-4, 1e-4\}, and utilize a hybrid time-frequency MAE loss named Fredf \cite{wang2024fredf} to improve prediction stability.

        \noindent\textbf{Efficiency.}
        The dual-branch ASNA introduces only a constant-factor increase in attention computation, approximately from $O(C(L/S)^2)$ to $O(2C(L/S)^2)$, while temporal downsampling reduces the practical cost on high-dimensional datasets. Empirically, on ETTm2/Weather/ECL, SeesawNet reaches 0.16/0.02/25.16 GFLOPs and 1.50/0.95/5.91 ms latency, remaining comparable to strong baselines.

    \subsection{Main Experiments}
    \begin{table*}[!ht]
        \centering
        \fontsize{8}{9.5}\selectfont
        \setlength{\tabcolsep}{4.0pt}

        \begin{tabularx}{1\textwidth}{c@{\hspace{7pt}}|c|c@{\hspace{3pt}}c|c@{\hspace{3pt}}c|c@{\hspace{3pt}}c|c@{\hspace{3pt}}c|c@{\hspace{3pt}}c|c@{\hspace{3pt}}c|c@{\hspace{3pt}}c|c@{\hspace{3pt}}c|c@{\hspace{3pt}}c}

        \toprule
        \multicolumn{2}{c|}{Models}
            & \multicolumn{2}{c|}{\makecell{SeesawNet \\ (Ours)}}
            & \multicolumn{2}{c|}{\makecell{DDN \\ (2024)}}
            & \multicolumn{2}{c|}{\makecell{SAN \\ (2023)}}
            & \multicolumn{2}{c|}{\makecell{TimeBridge \\ (2025)}}
            & \multicolumn{2}{c|}{\makecell{U-mixer \\ (2024)}}
            & \multicolumn{2}{c|}{\makecell{NS-Transformer \\ (2022)}}
            & \multicolumn{2}{c|}{\makecell{PDF \\ (2024)}}
            & \multicolumn{2}{c|}{\makecell{iTransformer \\ (2024)}}
            & \multicolumn{2}{c}{\makecell{PatchTST \\ (2023)}} \\
        \midrule

        \multicolumn{2}{c|}{Metric}
            & \textbf{MSE} & \textbf{MAE}
            & \textbf{MSE} & \textbf{MAE}
            & \textbf{MSE} & \textbf{MAE}
            & \textbf{MSE} & \textbf{MAE}
            & \textbf{MSE} & \textbf{MAE}
            & \textbf{MSE} & \textbf{MAE}
            & \textbf{MSE} & \textbf{MAE}
            & \textbf{MSE} & \textbf{MAE}
            & \textbf{MSE} & \textbf{MAE}\\
        \midrule

        \multirow{4}{*}{\rotatebox{90}{ETTh1}}
            & 96 & \textbf{0.353} & \textbf{0.389} & 0.379 & 0.405 & 0.383 & 0.399 & \underline{0.354} & \underline{0.391} & 0.370 & \textbf{0.389} & 0.547 & 0.501 & 0.362 & 0.394 & 0.402 & 0.418 & 0.377 & 0.401 \\
            & 192 & \textbf{0.387} & \textbf{0.410} & 0.419 & 0.432 & 0.419 & 0.419 & \underline{0.391} & \underline{0.416} & 0.418 & 0.419 & 0.603 & 0.529 & 0.394 & 0.416 & 0.449 & 0.448 & 0.414 & 0.421 \\
            & 336 & \textbf{0.390} & \textbf{0.421} & 0.462 & 0.460 & 0.438 & \underline{0.432} & \underline{0.415} & 0.434 & 0.465 & 0.444 & 0.711 & 0.585 & 0.422 & 0.434 & 0.474 & 0.466 & 0.428 & 0.434 \\
            & 720 & \textbf{0.434} & \textbf{0.460} & 0.562 & 0.531 & 0.447 & \textbf{0.460} & \underline{0.443} & \underline{0.462} & 0.526 & 0.497 & 0.714 & 0.601 & 0.469 & 0.482 & 0.569 & 0.540 & 0.447 & 0.464 \\
        \midrule

        \multirow{4}{*}{\rotatebox{90}{ETTh2}}
            & 96 & \textbf{0.258} & \textbf{0.322} & 0.279 & 0.342 & 0.277 & 0.338 & \underline{0.271} & \underline{0.332} & 0.291 & 0.336 & 0.392 & 0.420 & 0.273 & 0.335 & 0.302 & 0.359 & 0.275 & 0.336 \\
            & 192 & \textbf{0.314} & \textbf{0.361} & 0.341 & 0.384 & 0.340 & 0.378 & 0.339 & 0.376 & 0.386 & 0.395 & 0.519 & 0.480 & \underline{0.335} & \underline{0.376} & 0.390 & 0.412 & 0.340 & 0.379 \\
            & 336 & \textbf{0.314} & \textbf{0.371} & 0.369 & 0.410 & 0.362 & 0.401 & 0.370 & 0.403 & 0.441 & 0.430 & 0.572 & 0.511 & 0.360 & 0.401 & 0.426 & 0.438 & \underline{0.329} & \underline{0.382} \\
            & 720 & \textbf{0.372} & \textbf{0.416} & 0.406 & 0.447 & 0.398 & 0.436 & 0.420 & 0.447 & 0.487 & 0.465 & 0.600 & 0.536 & 0.393 & 0.434 & 0.421 & 0.446 & \underline{0.379} & \underline{0.421} \\
        \midrule

        \multirow{4}{*}{\rotatebox{90}{ETTm1}}
            & 96 & 0.293 & 0.344 & 0.313 & 0.356 & 0.288 & 0.342 & \underline{0.279} & \textbf{0.333} & 0.321 & 0.351 & 0.393 & 0.404 & \textbf{0.278} & \underline{0.338} & 0.303 & 0.357 & 0.291 & 0.342 \\
            & 192 & \textbf{0.315} & \textbf{0.361} & 0.343 & 0.373 & 0.323 & 0.363 & 0.321 & 0.363 & 0.361 & 0.371 & 0.481 & 0.447 & \underline{0.316} & \underline{0.362} & 0.344 & 0.381 & 0.332 & 0.369 \\
            & 336 & \textbf{0.342} & \textbf{0.379} & 0.386 & 0.397 & 0.357 & 0.384 & 0.358 & 0.388 & 0.395 & 0.395 & 0.555 & 0.489 & \underline{0.349} & \underline{0.381} & 0.380 & 0.404 & 0.367 & 0.392 \\
            & 720 & \textbf{0.398} & \textbf{0.407} & 0.433 & 0.424 & \underline{0.409} & 0.415 & 0.416 & 0.418 & 0.459 & 0.432 & 0.677 & 0.548 & 0.415 & \underline{0.413} & 0.441 & 0.438 & 0.415 & 0.422 \\
        \midrule

        \multirow{4}{*}{\rotatebox{90}{ETTm2}}
            & 96 & \underline{0.160} & \underline{0.244} & 0.162 & 0.253 & 0.165 & 0.257 & \textbf{0.157} & \textbf{0.243} & 0.176 & 0.256 & 0.266 & 0.324 & 0.162 & 0.254 & 0.177 & 0.269 & 0.164 & 0.253 \\
            & 192 & \textbf{0.215} & \textbf{0.284} & \underline{0.217} & \underline{0.291} & 0.221 & 0.295 & \underline{0.217} & \textbf{0.284} & 0.244 & 0.301 & 0.434 & 0.405 & \underline{0.217} & 0.292 & 0.243 & 0.313 & 0.221 & 0.293 \\
            & 336 & \textbf{0.267} & \textbf{0.320} & 0.269 & \underline{0.327} & \underline{0.268} & 0.328 & 0.270 & \textbf{0.320} & 0.310 & 0.343 & 0.808 & 0.563 & \underline{0.268} & \underline{0.327} & 0.291 & 0.343 & 0.276 & 0.329 \\
            & 720 & \underline{0.350} & \textbf{0.373} & 0.350 & 0.380 & 0.358 & 0.384 & \underline{0.350} & \underline{0.375} & 0.447 & 0.418 & 0.758 & 0.555 & \textbf{0.347} & 0.378 & 0.377 & 0.396 & 0.367 & 0.385 \\
        \midrule

        \multirow{4}{*}{\rotatebox{90}{Exchange}}
            & 96 & \underline{0.088} & \underline{0.213} & 0.094 & 0.217 & \textbf{0.087} & 0.214 & 0.091 & 0.215 & \textbf{0.087} & \textbf{0.201} & 0.132 & 0.254 & 0.094 & 0.219 & 0.099 & 0.226 & 0.091 & 0.214 \\
            & 192 & 0.181 & \underline{0.311} & 0.193 & 0.314 & \underline{0.177} & 0.316 & 0.198 & 0.318 & \textbf{0.171} & \textbf{0.295} & 0.258 & 0.363 & 0.198 & 0.319 & 0.206 & 0.329 & 0.207 & 0.326 \\
            & 336 & \textbf{0.289} & \textbf{0.397} & 0.350 & 0.427 & \underline{0.296} & 0.410 & 0.394 & 0.460 & 0.310 & \underline{0.404} & 0.470 & 0.494 & 0.332 & 0.420 & 0.379 & 0.456 & 0.346 & 0.429 \\
            & 720 & \underline{0.613} & \underline{0.602} & 0.999 & 0.751 & 0.714 & 0.642 & 1.151 & 0.820 & \textbf{0.561} & \textbf{0.566} & 1.518 & 0.896 & 0.950 & 0.727 & 0.953 & 0.743 & 0.895 & 0.711 \\

        \midrule

        \multirow{4}{*}{\rotatebox{90}{Weather}}
            & 96 & \textbf{0.142} & \textbf{0.184} & 0.153 & 0.211 & 0.152 & 0.210 & \underline{0.144} & \underline{0.185} & 0.159 & 0.197 & 0.183 & 0.230 & 0.145 & 0.196 & 0.163 & 0.214 & 0.150 & 0.199 \\
            & 192 & 0.190 & \underline{0.235} & 0.200 & 0.259 & 0.196 & 0.253 & \textbf{0.186} & \textbf{0.227} & 0.203 & 0.239 & 0.253 & 0.291 & \underline{0.189} & 0.239 & 0.204 & 0.250 & 0.195 & 0.241 \\
            & 336 & \textbf{0.238} & \underline{0.272} & 0.248 & 0.300 & 0.246 & 0.294 & \textbf{0.238} & \textbf{0.269} & 0.253 & 0.278 & 0.326 & 0.341 & \underline{0.243} & 0.281 & 0.255 & 0.289 & 0.248 & 0.282 \\
            & 720 & \textbf{0.289} & \textbf{0.314} & 0.317 & 0.350 & 0.315 & 0.346 & 0.312 & \underline{0.325} & 0.329 & 0.331 & 0.417 & 0.397 & \underline{0.310} & 0.330 & 0.328 & 0.339 & 0.322 & 0.335 \\
        \midrule

        \multirow{4}{*}{\rotatebox{90}{ILI}}
            & 24 & \textbf{1.562} & \textbf{0.808} & 1.830 & 0.950 & 2.297 & 1.055 & 1.938 & \underline{0.857} & 2.398 & 0.972 & 2.420 & 0.959 & \underline{1.829} & \underline{0.857} & 2.032 & 0.951 & 2.160 & 0.928 \\
            & 36 & \textbf{1.658} & \textbf{0.829} & 2.142 & 1.000 & 2.323 & 1.070 & \underline{1.772} & \underline{0.857} & 2.606 & 1.005 & 3.093 & 1.078 & 2.008 & 0.960 & 1.984 & 0.959 & 1.902 & 0.910 \\
            & 48 & \textbf{1.567} & \textbf{0.828} & 2.129 & 1.000 & 2.262 & 1.065 & \underline{1.702} & \underline{0.851} & 2.495 & 0.994 & 2.500 & 0.990 & 2.224 & 1.020 & 2.062 & 0.994 & 1.734 & 0.881 \\
            & 60 & \textbf{1.702} & \textbf{0.879} & 2.230 & 1.025 & 2.443 & 1.124 & \underline{1.733} & \underline{0.887} & 2.609 & 1.022 & 2.378 & 0.999 & 2.147 & 1.000 & 2.117 & 1.019 & 1.861 & 0.932 \\
        \midrule

        \multirow{4}{*}{\rotatebox{90}{Solar}}
            & 96 & 0.158 & 0.201 & 0.180 & 0.230 & \textbf{0.139} & \underline{0.192} & 0.163 & 0.216 & 0.187 & 0.236 & \underline{0.154} & \textbf{0.177} & 0.185 & 0.255 & 0.205 & 0.258 & 0.184 & 0.243 \\
            & 192 & \textbf{0.171} & \textbf{0.216} & 0.199 & 0.253 & \underline{0.175} & 0.225 & 0.179 & 0.234 & 0.222 & 0.258 & 0.182 & \underline{0.217} & 0.216 & 0.270 & 0.230 & 0.269 & 0.190 & 0.244 \\
            & 336 & \textbf{0.182} & \textbf{0.223} & 0.203 & 0.250 & \underline{0.188} & 0.241 & 0.192 & 0.230 & 0.231 & 0.276 & 0.200 & \underline{0.225} & 0.230 & 0.282 & 0.229 & 0.272 & 0.204 & 0.261 \\
            & 720 & \textbf{0.193} & \textbf{0.232} & 0.218 & 0.261 & 0.203 & 0.250 & \underline{0.203} & 0.251 & 0.238 & 0.281 & \underline{0.203} & \underline{0.239} & 0.236 & 0.296 & 0.226 & 0.277 & 0.211 & 0.258 \\
        \midrule

        \multirow{4}{*}{\rotatebox{90}{ECL}}
            & 96 & 0.128 & \underline{0.221} & 0.127 & 0.224 & 0.137 & 0.234 & \textbf{0.121} & \textbf{0.216} & 0.168 & 0.263 & 0.172 & 0.276 & \underline{0.127} & \underline{0.221} & 0.132 & 0.227 & 0.131 & 0.225 \\
            & 192 & \textbf{0.144} & \textbf{0.237} & 0.147 & 0.245 & 0.152 & 0.248 & \textbf{0.144} & \underline{0.240} & 0.177 & 0.268 & 0.185 & 0.287 & \underline{0.146} & \underline{0.240} & 0.155 & 0.249 & 0.149 & 0.242 \\
            & 336 & \underline{0.161} & \textbf{0.253} & \textbf{0.154} & \underline{0.254} & 0.167 & 0.264 & 0.162 & 0.257 & 0.190 & 0.283 & 0.196 & 0.300 & 0.162 & 0.257 & 0.170 & 0.265 & 0.168 & 0.263 \\
            & 720 & 0.182 & 0.283 & \textbf{0.177} & \underline{0.279} & 0.202 & 0.296 & \underline{0.179} & \textbf{0.273} & 0.227 & 0.316 & 0.221 & 0.319 & 0.196 & 0.288 & 0.204 & 0.297 & 0.208 & 0.297 \\
        \midrule

        \multicolumn{2}{c|}{Improvement}
         & - & - & 9.7\% & 7.5\% & 7.2\% & 6.0\% & 5.3\% & 3.0\% & 15.5\% & 7.7\% & 32.7\% & 19.2\% & 8.1\% & 6.2\% & 14.5\% & 10.2\% & 7.9\% & 5.3\% \\

        \bottomrule
        \end{tabularx}

        \caption{Long-term forecasting performance comparison. All results are tested across four different prediction lengths $H \in \{96, 192, 336, 720\}$, except that ILI uses $H \in \{24, 36, 48, 60\}$. The best results are in \textbf{bold} and the second best are \underline{underlined}.
        }
        \label{tab1}
        \end{table*}

        \noindent\textbf{Long-term Forecasting.}
            Table \ref{tab1} reports MSE/MAE over nine datasets and four horizons. SeesawNet achieves the best result in 51 of 72 settings and ranks second in 12, showing consistent competitiveness across diverse non-stationary scenarios. Compared with IN-based models such as DDN and SAN, SeesawNet reduces MSE/MAE by 8.5\%/6.8\% on average, with particularly clear gains on ETT where instance-specific temporal structures are rich. Against dependency-modeling baselines such as U-Mixer, NS-Transformer, and TimeBridge, the gains increase to 18.1\%/9.2\%; on highly non-stationary datasets such as ILI and Solar, the adaptive balance between common and specific dependencies is especially beneficial. Compared with strong sequence forecasters such as PDF, PatchTST, and iTransformer, SeesawNet still reduces MSE/MAE by 10.2\%/7.2\%. The gains are smaller on Exchange/ECL because weak cross-variable cointegration or strong shared load patterns reduce the relative benefit of instance-specific channel modeling.

        \refstepcounter{table}\label{tab2}
        \noindent\textbf{ASNA Generality.}
            To evaluate generality, we replace the attention backbone with ASNA in PatchTST (channel-independent temporal patch modeling) and iTransformer (channel-dependent inverted tokens), with corresponding input adaptations. As shown in Table \ref{tab2}, ASNA consistently improves both models: iTransformer gains 7.7\%/5.6\% on average, while PatchTST reduces MSE/MAE by 3.8\%/4.0\%.

    \subsection{Ablation Study}

        \begin{figure*}[t]
            \centering
            \includegraphics[width= 0.85\linewidth]{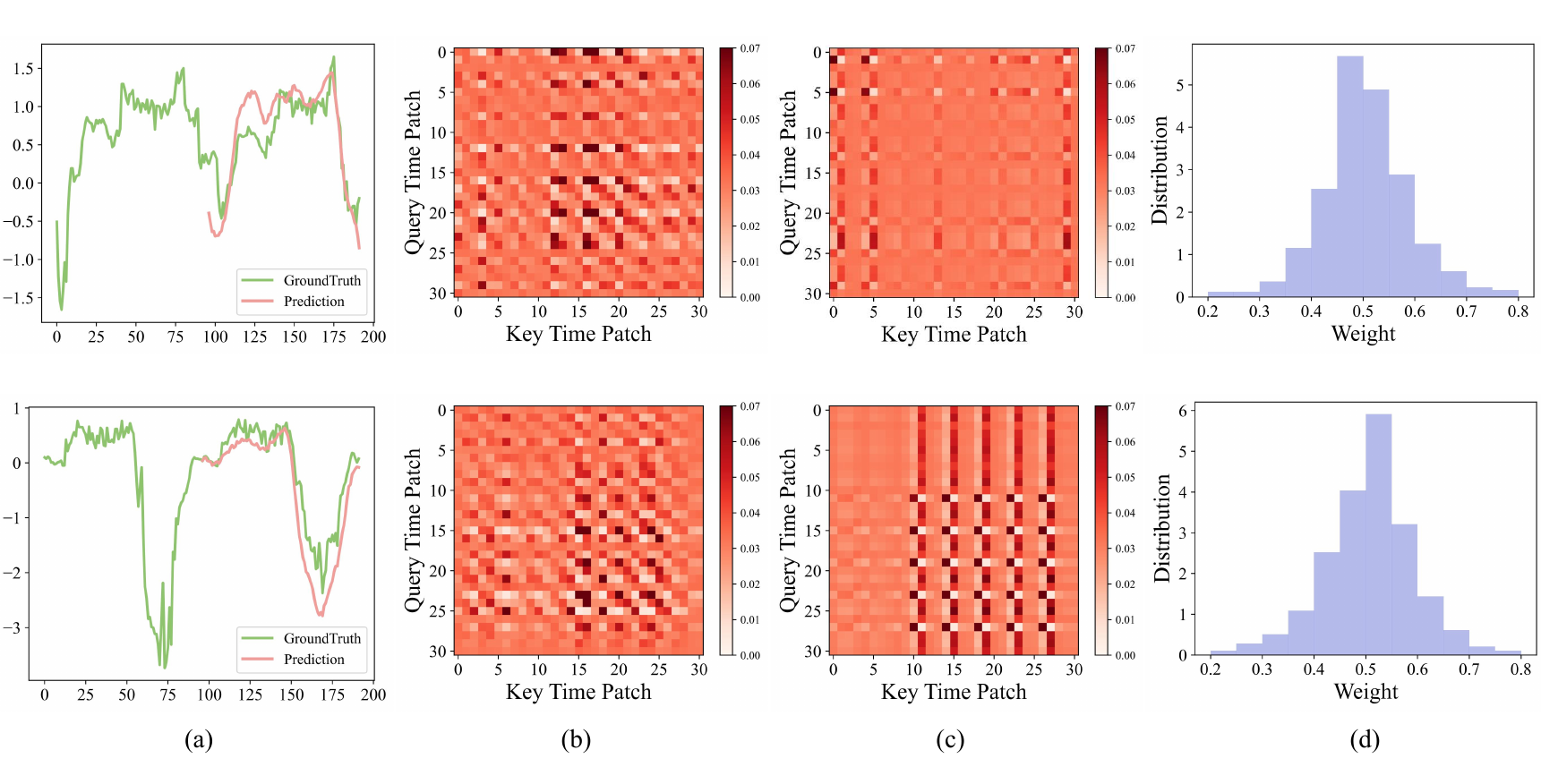}
            \caption{Visualization of SeesawNet’s attention behavior on two samples from ETTm1. Each row presents one sample. (a) shows the ground truth and predicted sequences. (b) and (c) display the stationary and non-stationary attention heatmaps, respectively. (d) illustrates the dynamic weight distributions assigned to stationary attention by the adaptive gating mechanism.}
            \label{fig4}
        \end{figure*}

        \begin{table}[hbt]
            \centering
            \fontsize{8}{9.5}\selectfont
            \setlength{\tabcolsep}{4.5pt}
            \begin{tabularx}{0.99\linewidth}{@{}>{\centering\arraybackslash}m{0.18\linewidth}|>{\centering\arraybackslash}X@{\hspace{3pt}}>{\centering\arraybackslash}X|>{\centering\arraybackslash}X@{\hspace{3pt}}>{\centering\arraybackslash}X|>{\centering\arraybackslash}X@{\hspace{3pt}}>{\centering\arraybackslash}X|>{\centering\arraybackslash}X@{\hspace{3pt}}>{\centering\arraybackslash}X@{}}
            \toprule
            \multirow{2}{*}{\centering Variant}
                & \multicolumn{2}{c|}{96}
                & \multicolumn{2}{c|}{192}
                & \multicolumn{2}{c|}{336}
                & \multicolumn{2}{c}{720} \\
            \cmidrule(lr){2-3}\cmidrule(lr){4-5}\cmidrule(lr){6-7}\cmidrule(l){8-9}
                & MSE & MAE & MSE & MAE & MSE & MAE & MSE & MAE \\
            \midrule
            Full-Model
                & \textbf{0.198} & 0.265 & \textbf{0.240} & 0.302
                & \textbf{0.273} & \textbf{0.330} & \textbf{0.355} & \textbf{0.386} \\
            \midrule
            \textbf{\textit{w/o sta}}
                & 0.227 & 0.301 & 0.270 & 0.331 & 0.336 & 0.372 & 0.506 & 0.459 \\
            \textbf{\textit{w/o non}}
                & 0.204 & \textbf{0.264} & 0.248 & \textbf{0.300} & 0.289 & 0.331 & 0.406 & 0.407 \\
            \textbf{\textit{w/o gate}}
                & 0.200 & 0.269 & 0.243 & 0.304 & 0.276 & 0.333 & 0.367 & 0.393 \\
            \midrule
            \textbf{\textit{w/o pd}}
                & 0.206 & 0.269 & 0.250 & 0.305 & 0.291 & 0.337 & 0.406 & 0.404 \\
            \textbf{\textit{w/o cr}}
                & 0.201 & 0.268 & 0.243 & 0.304 & 0.276 & 0.332 & 0.361 & 0.390 \\
            \textbf{\textit{cr+pd}}
                & 0.203 & 0.269 & 0.248 & 0.306 & 0.287 & 0.335 & 0.400 & 0.403 \\
            \bottomrule
            \end{tabularx}
            \caption{Average ablation results over eight datasets.}
            \label{tab:ablation_avg}
        \end{table}

        \noindent\textbf{Common vs. Specific Dependencies.}
            Table \ref{tab:ablation_avg} ablates ASNA components: removing stationary attention (\textbf{\textit{w/o sta}}), non-stationary attention (\textbf{\textit{w/o non}}), or gating (\textbf{\textit{w/o gate}}) degrades performance by 18.6\%, 8.1\%, and 1.5\% on average, respectively, confirming that both branches are necessary and the gate provides adaptive coordination.

        \begin{center}
            \fontsize{8}{9.5}\selectfont
            \setlength{\tabcolsep}{4.5pt}
            \begin{tabularx}{0.99\linewidth}{c|c@{\hspace{3pt}}c|c@{\hspace{3pt}}c|c@{\hspace{3pt}}c|c@{\hspace{3pt}}c}
            \toprule
            Model &
                \multicolumn{4}{c|}{iTransformer} & \multicolumn{4}{c}{PatchTST} \\
            \midrule
            Backbone &
                \multicolumn{2}{c|}{Transformer} & \multicolumn{2}{c|}{ASNA} &
                \multicolumn{2}{c|}{Transformer} & \multicolumn{2}{c}{ASNA} \\
            \midrule
            Metric
                & \textbf{MSE} & \textbf{MAE}
                & \textbf{MSE} & \textbf{MAE}
                & \textbf{MSE} & \textbf{MAE}
                & \textbf{MSE} & \textbf{MAE}\\ \midrule
            ETTh1 & 0.474 & 0.468 & \textbf{0.438} & \textbf{0.444} & 0.417 & 0.430 & \textbf{0.405} & \textbf{0.420} \\
            ETTh2 & 0.385 & 0.414 & \textbf{0.366} & \textbf{0.400} & 0.331 & 0.380 & \textbf{0.325} & \textbf{0.377} \\
            ETTM1 & 0.367 & 0.395 & \textbf{0.355} & \textbf{0.382} & 0.351 & 0.381 & \textbf{0.345} & \textbf{0.374} \\
            ETTM2 & 0.272 & 0.330 & \textbf{0.264} & \textbf{0.322} & 0.257 & 0.315 & \textbf{0.253} & \textbf{0.309} \\
            Exchange & 0.409 & 0.438 & \textbf{0.313} & \textbf{0.396} & 0.385 & 0.420 & \textbf{0.330} & \textbf{0.400} \\
            Weather & 0.238 & 0.273 & \textbf{0.226} & \textbf{0.261} & 0.229 & 0.264 & \textbf{0.222} & \textbf{0.251} \\
            Solar & 0.222 & 0.269 & \textbf{0.208} & \textbf{0.240} & 0.197 & 0.251 & \textbf{0.195} & \textbf{0.222} \\
            \midrule
            Promotion & - & - & 7.7\% & 5.6\% & - & - & 3.8\% & 4.0\% \\
            \bottomrule
            \end{tabularx}
            \captionof*{table}{Table~\ref{tab2}: The generalization capability of ASNA as a decoupled and universal backbone.}
        \end{center}

    \noindent\textbf{Temporal vs. Cross-channel Dependency.}
        Table \ref{tab:ablation_avg} also reports four variants: SeesawNet, \textbf{\textit{w/o cr}}, \textbf{\textit{w/o pd}}, and order reversal (\textbf{\textit{cr+pd}}). Removing patch modeling and channel modeling degrades average performance by 8.1\% and 3.5\%, respectively, showing that the full temporal-then-channel design provides the best overall trade-off.

        \noindent\textbf{Effect of Training Objective (Fredf vs. MSE).}
            Our main experiments adopt the Fredf loss, which suppresses the auto-correlation of prediction errors in the frequency domain and has been shown to improve forecasting accuracy.
            To verify that the gains of SeesawNet
            , we conduct an ablation by replacing Fredf with the standard MSE loss.

            As reported in Table~\ref{tab:alb_loss}, using MSE slightly degrades SeesawNet: the average MSE increases by $0.1\%$ and the average MAE increases by $0.9\%$ across datasets.
            Nevertheless, SeesawNet trained with MSE still outperforms competing methods by a clear margin (a relative improvement of $4.7\%$--$8.9\%$ on average), indicating that the superiority of SeesawNet primarily comes from the proposed balancing architecture rather than being dominated by the loss design.

    \subsection{Visualization}
        \noindent\textbf{Attention Heatmap Visualization.}
            Figure \ref{fig4} visualizes stationary/non-stationary attention and adaptive gate weights on two ETTm1 samples. Despite different raw distributions, stationary maps remain similar across samples, indicating shared patterns learned from normalized inputs; in contrast, non-stationary maps vary substantially, reflecting instance-specific structures preserved from raw sequences. The sample farther from the distribution center receives higher stationary weight from the gate, suggesting that SeesawNet adaptively relies more on common dependencies when uncertainty is larger. Overall, the visualization supports that SeesawNet separates and balances common and specific dependencies.

\begin{table}[!hbt]
    \centering
    \fontsize{8}{9.5}\selectfont
    \setlength{\tabcolsep}{4.5pt}

    \setlength{\tabcolsep}{2.8pt}
    \renewcommand{\arraystretch}{1.0}
    \setlength{\arrayrulewidth}{0.35pt}

    \begin{tabularx}{0.99\linewidth}{@{}
    c|c|>{\centering\arraybackslash}X|>{\centering\arraybackslash}X|
    >{\centering\arraybackslash}X|>{\centering\arraybackslash}X|
    >{\centering\arraybackslash}X|>{\centering\arraybackslash}X@{}}
    \toprule
    \multirow{2}{*}{Dataset} &
    Model &
    \multicolumn{2}{c|}{SeesawNet} &
    {\scriptsize TimeBr.} & SAN & {\scriptsize U-Mixer} & PDF \\
    \cmidrule{2-8}
    & Loss & Fredf & MSE & MSE & hybrid & MAE & MSE \\
    \midrule

    \multirow{2}{*}{\centering ETTh1}
        & MSE & \textbf{0.391} & \underline{0.398} & 0.422 & 0.422 & 0.445 & 0.412 \\
        & MAE & \textbf{0.420} & \underline{0.425} & 0.443 & 0.428 & 0.437 & 0.431 \\
    \midrule

    \multirow{2}{*}{\centering ETTh2}
        & MSE & \underline{0.314} & \textbf{0.309} & 0.362 & 0.344 & 0.401 & 0.340 \\
        & MAE & \textbf{0.368} & \underline{0.369} & 0.403 & 0.388 & 0.407 & 0.387 \\
    \midrule

    \multirow{2}{*}{\centering ETTm1}
        & MSE & \textbf{0.337} & 0.345 & 0.357 & 0.345 & 0.384 & \underline{0.340} \\
        & MAE & \textbf{0.373} & 0.377 & 0.392 & 0.376 & 0.387 & \underline{0.374} \\
    \midrule

    \multirow{2}{*}{\centering ETTm2}
        & MSE & \textbf{0.248} & \textbf{0.248} & 0.266 & 0.253 & 0.294 & \underline{0.249} \\
        & MAE & \textbf{0.305} & \underline{0.311} & 0.328 & 0.316 & 0.329 & 0.313 \\
    \midrule

    \multirow{2}{*}{\centering \scriptsize Exchange}
        & MSE & 0.293 & \underline{0.284} & 0.384 & 0.318 & \textbf{0.282} & 0.394 \\
        & MAE & \underline{0.381} & 0.383 & 0.410 & 0.395 & \textbf{0.366} & 0.421 \\
    \midrule

    \multirow{2}{*}{\centering Weather}
        & MSE & \textbf{0.215} & \textbf{0.215} & \underline{0.222} & 0.227 & 0.236 & \underline{0.222} \\
        & MAE & \textbf{0.251} & \underline{0.252} & 0.263 & 0.276 & 0.261 & 0.261 \\
    \bottomrule
    \end{tabularx}

    \caption{Ablation on training objectives. }
    \label{tab:alb_loss}
\end{table}

\section{Conclusion}
    In this paper, we propose SeesawNet, an ASNA-based framework for balancing common and instance-specific dependencies in non-stationary time series forecasting. ASNA learns shared patterns from normalized inputs and complementary instance-specific structures from raw sequences, then adaptively fuses them according to each instance's non-stationarity. Built on this mechanism, SeesawNet extends the balance to both temporal patch modeling and cross-channel relationship learning. Experiments, generalization studies, ablations, and visualizations show consistent gains and demonstrate that ASNA can strengthen other backbones. Future work will study how different non-stationary patterns affect this balance.

\clearpage

\section*{Acknowledgements}
The work of H. Li, C. Liu, and Y. Zhou was supported in part by the National Natural Science Foundation of China (NSFC) under Grant No. 62171302.
The work of P. Wang was supported in part by the Science and Technology Development Fund, Macao SAR (No. 001/2024/SKL and No. 0002/2025/EQP, No.0072/2025/AMJ), the University of Macau (No. MYRG-GRG2025-00241-IOTSC).

\bibliographystyle{named}
\bibliography{ijcai26}

\end{document}